\documentclass[10pt,twocolumn,letterpaper]{article}

\usepackage{cvpr}               % To produce the CAMERA-READY version

\usepackage{graphicx}
\usepackage{capt-of}
\usepackage{xcolor}
\usepackage{amsmath,amssymb,amsfonts,dsfont,pifont,bm,bbm,mathrsfs,mathtools,nicefrac}
\usepackage{algorithm,algpseudocode,listings}
\usepackage{bm}
\usepackage{multirow}
\usepackage{booktabs,multirow,adjustbox,diagbox,threeparttable}

\usepackage{array}

\definecolor{citeblue}{RGB}{48,111,186}
\usepackage[pagebackref,breaklinks,colorlinks,citecolor=citeblue,bookmarks=false]{hyperref}
\usepackage[capitalize]{cleveref}  % Should be loaded after 'hyperref', and works perfectly with 'subfigure'.

\crefname{section}{Sec.}{Secs.}
\Crefname{section}{Section}{Sections}
\crefname{table}{Tab.}{Tabs.}
\Crefname{table}{Table}{Tables}
\crefname{figure}{Fig.}{Figs.}
\Crefname{figure}{Figure}{Figures}
\crefname{equation}{Eq.}{Eqs.}
\Crefname{equation}{Equation}{Equations}
\newcommand{\tablestyle}[2]{\setlength{\tabcolsep}{#1}\renewcommand{\arraystretch}{#2}\centering\small}
  
\newlength\savewidth\newcommand\shline{\noalign{\global\savewidth\arrayrulewidth
  \global\arrayrulewidth 1pt}\hline\noalign{\global\arrayrulewidth\savewidth}}

\def\cvprPaperID{1365}
\def\confName{CVPR}
\def\confYear{2023}

\begin{document}

\title{I2VGen-XL: High-Quality Image-to-Video Synthesis \\ via Cascaded Diffusion Models}

\author{
  Shiwei Zhang$^{1*}$ \quad Jiayu Wang$^{1*}$ \quad Yingya Zhang$^{1*}$ \quad Kang Zhao$^1$ \quad Hangjie Yuan$^2$ \\ Zhiwu Qin$^3$ \quad  Xiang Wang$^3$  \quad Deli Zhao$^1$ \quad  Jingren Zhou$^1$\\
    % \vspace{-2mm}
    $^1$Alibaba Group \quad  $^2$ Zhejiang University \\
    $^3$ Huazhong University of Science and Technology \\
    {\tt\small \{zhangjin.zsw, wangjiayu.wjy\}@alibaba-inc.com} \\
    {\tt\small \{yingya.zyy, zhaokang.zk, jingren.zhou\}@alibaba-inc.com} \\
    {\tt\small hj.yuan@zju.edu.cn \quad \{qzw, wxiang\}@hust.edu.cn \quad zhaodeli@gmail.com} \\
    \vspace{-0.6cm}
}

% \maketitle
\twocolumn[{%
\maketitle
\vspace{-2.0em}
\begin{center}
\centering
\vspace{-5pt}
\includegraphics[width=1.0\linewidth]{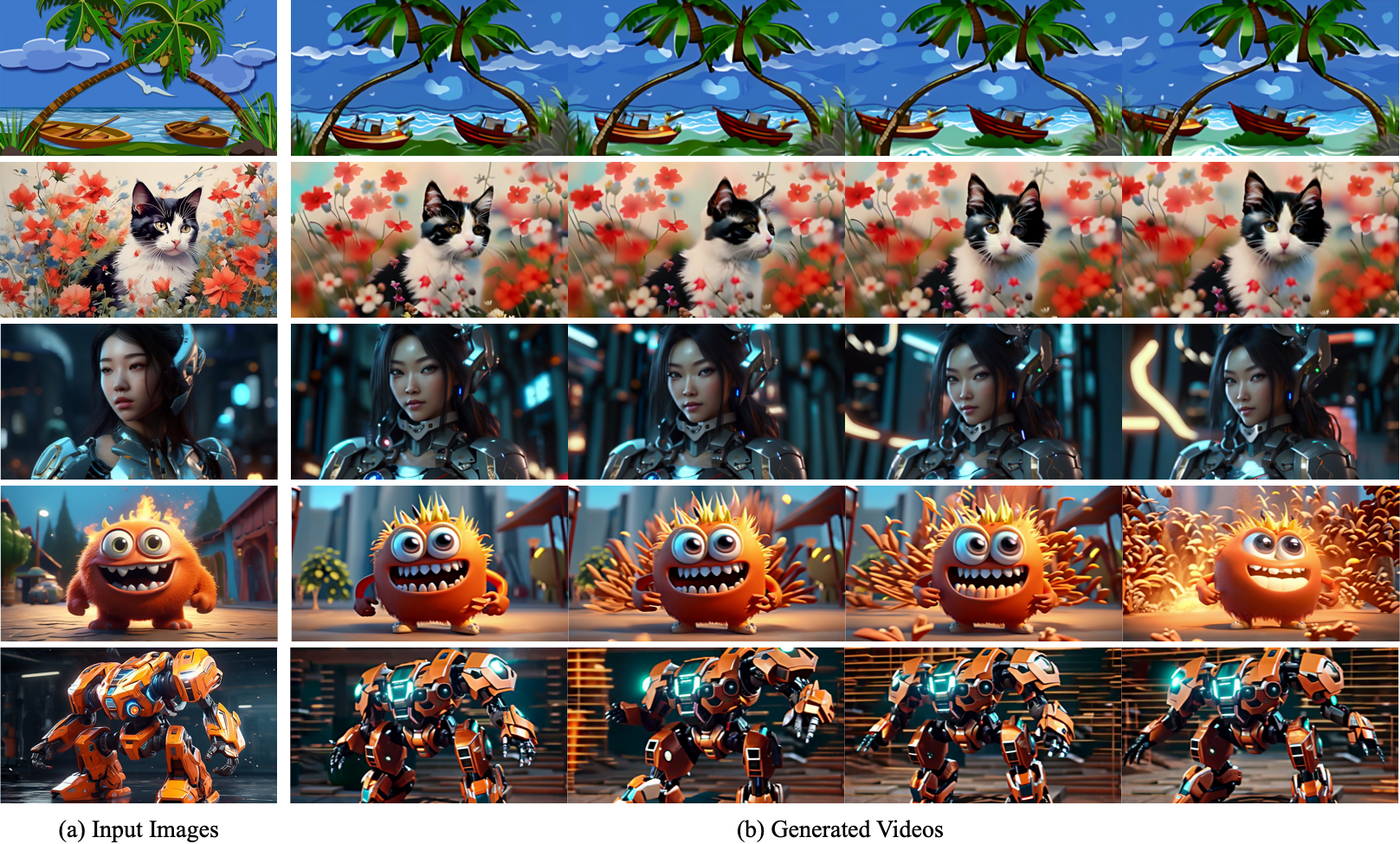}
\vspace{-18pt}
\captionsetup{type=figure}
\caption{
    Examples of the generated videos. 
    I2VGen-XL is capable of generating high-quality videos of various categories, such as art, humans, animals, technology, and more. 
    The generated videos exhibit advantages like high definition, high resolution, smoothness, and aesthetic appeal, catering to a wider range of video content creation.
}
\label{fig:teaser}
\vspace{-3mm}
\end{center}
}]

\newcommand\blfootnote[1]{%
\begingroup
\renewcommand\thefootnote{}\footnote{#1}%
\addtocounter{footnote}{-1}%
\endgroup
}

\blfootnote{$^*$ Equal contribution.}

%%%%%%%%% ABSTRACT
\begin{abstract}
\vspace{-0.2cm}
Video synthesis has recently made remarkable strides benefiting from the rapid development of diffusion models.
However, it still encounters challenges in terms of semantic accuracy, clarity and spatio-temporal continuity.
They primarily arise from the scarcity of well-aligned text-video data and the complex inherent structure of videos, making it difficult for the model to simultaneously ensure semantic and qualitative excellence.
In this report, we propose a cascaded I2VGen-XL approach that enhances model performance by decoupling these two factors and ensures the alignment of the input data by utilizing static images as a form of crucial guidance.
I2VGen-XL consists of two stages: 
i) the base stage guarantees coherent semantics and preserves content from input images by using two hierarchical encoders, and 
ii) the refinement stage enhances the video's details by incorporating an additional brief text and improves the resolution to 1280$\times$720.
To improve the diversity, we collect around 35 million single-shot text-video pairs and 6 billion text-image pairs to optimize the model.
By this means, I2VGen-XL can simultaneously enhance the semantic accuracy, continuity of details and clarity of generated videos.
Through extensive experiments, we have investigated the underlying principles of I2VGen-XL and compared it with current top methods, which can demonstrate its effectiveness on diverse data.
The source code and models will be publicly available at \url{https://i2vgen-xl.github.io}.

\vspace{-5mm}
\end{abstract}
\section{Introduction}
\label{sec:intro}
Recently, the technological revolution brought by diffusion models~\cite{sohl2015Diffusion_model,ho2020denoising_ddpm} in image synthesis~\cite{rombach2022LDM} has been impressively remarkable, and it has also led to significant advancements in video synthesis~\cite{blattmann2023align_latents,singer2022make-a-video, luo2023videofusion}.
Typically, they can generate realistic videos with unprecedented levels of fidelity and diversity from a textual prompt as input~\cite{blattmann2023align_latents}, or even control content and motion patterns in videos based on different guidances~\cite{wang2023videocomposer,wang2023facecomposer}.
Despite these advancements, ensuring coherent semantics in both spatial and motion dimensions, as well as maintaining continuity of details in generated videos continues to pose significant challenges, thereby limiting its potential applications.

Current existing approaches to tackle this problem mainly fall into two categories.
The first category employs multiple models to progressively improve video quality like Imagen Video~\cite{ho2022imagenvideo}.
However, these methods primarily optimize the same objective using identical inputs at each stage, without explicitly decoupling the tasks. 
Consequently, this leads to the learning of similar distributions at each stage, resulting in non-negligible noise in generated videos.
The second category~\cite{wang2023videocomposer, ouyang2023codef} requires additional guidance or training processes.
While they have shown promising performance, fulfilling the guidance and training requirements in some scenarios remains a significant challenge.
Moreover, as a common challenge, the insufficient alignment of video-text pairs also significantly impedes the further advancement of video synthesis.

\begin{figure*}
    \centering
    \includegraphics[width=1.0\linewidth]{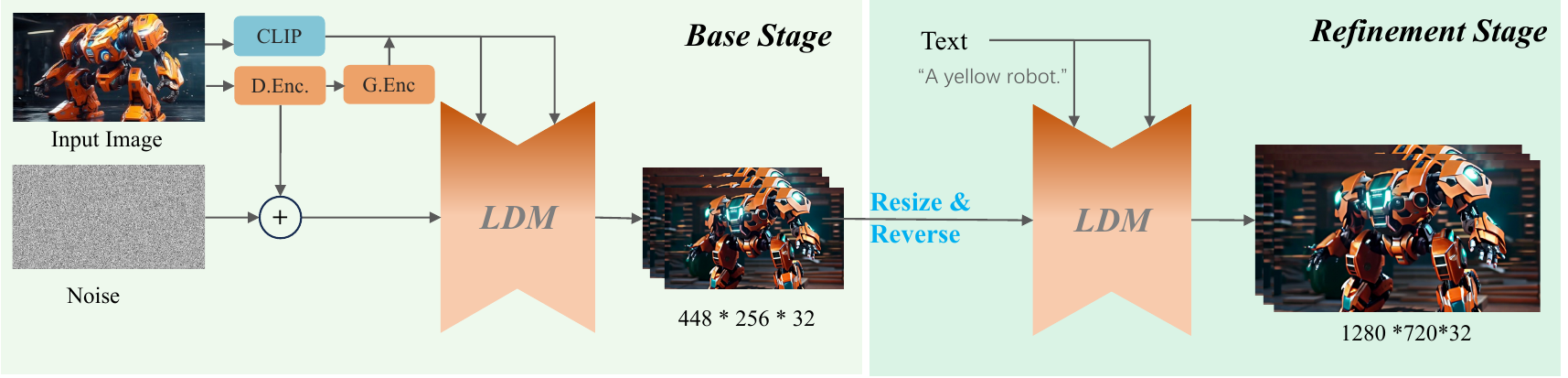}
    \caption{The overall framework of I2VGen-XL.
    In the \emph{base stage}, two hierarchical encoders are employed to simultaneously capture high-level semantics and low-level details of input images, ensuring more realistic dynamics while preserving the content and structure of images. 
    In the \emph{refinement stage}, a separate diffusion model is utilized to enhance the resolution and significantly improve the temporal continuity of videos by refining details.
    ``D.Enc.'' and ``G.Enc.'' denote the detail encoder and global encoder, respectively.
    }
    \label{fig:framrwork}
    \vspace{-4mm}
\end{figure*}

The success of SDXL~\cite{podell2023sdxl} inspires us to develop a cascaded I2VGen-XL method capable of generating high-definition videos with coherent spatial and motion dynamics, along with continuous details.
I2VGen-XL first reduces the reliance on well-aligned text-video pairs by utilizing a single static image as the primary condition, and it mainly consists of two stages as shown in Fig.~\ref{fig:framrwork}:
\emph{i) The base stage} aims to ensure semantic coherence in generated videos at a low resolution, while simultaneously preserving the content and identity information of input images.
For this purpose, we devise two hierarchical encoders, \emph{i.e.}, a fixed CLIP encoder and learnable content encoder, to respectively extract high-level semantics and low-level details, which are then incorporated into a video diffusion model.
\emph{ii) The refinement stage} is to improve the video resolution to 1280$\times$720 and refine details and artifacts that exist within generated videos. %resulting in a substantial improvement in clarity and overall quality.
Specifically, we train a distinct video diffusion model with a simple text as input, optimizing its initial 600 denoising steps.
By using the noising-denoising process~\cite{meng2021sdedit}, we can generate high-definition videos with both temporal and spatial coherence from low-resolution videos.

Furthermore, we have collected 35 million high-quality single-shot videos and 6 billion images, covering a broad spectrum of common categories in daily life, to enhance the diversity and robustness of I2VGen-XL.
Finally, through a comprehensive evaluation of a wide range of data, we meticulously analyze the effectiveness of I2VGen-XL. 
We delve into the working mechanism of the refinement model in the frequency domain and benchmark it against current top-ranked methods. 
The results demonstrate that I2VGen-XL exhibits more reasonable and significant motions in these cases.

\section{Related Works}
\noindent

\textbf{Diffusion probabilistic models.}
Diffusion Probabilistic Models (DPM)~\cite{sohl2015Diffusion_model} is a powerful class of generative models to learn a diffusion process that generates a target probability distribution.
The initial efforts in image generation utilizing DPMs were primarily centered around enhancing performance~\cite{kingma2021variational_diffusion_model,huang2023composer,dhariwal2021diffusion_beat_gan}. 
These models have been validated to be more effective than major traditional generation methods, including GANs~\cite{goodfellow2020GAN}  and VAE~\cite{kingma2013VAE}, in terms of diversity and realism. 
Therefore, DPMs have gradually emerged as a crucial branch in the field of generative methods, but they are plagued by inefficiency.
%
% Due to the inefficiency of DPM computations, a significant subsequent effort has been focused on improving efficiency.
%
For this purpose, some studies aim to reduce the number of denoising steps in the model by improving sampling efficiency, such as learning-free sampling~\cite{song2020denoising, song2020score_generative_SDE, liu2022pseudo_numerical, zhang2022fast_sampling_diffusion, zhang2022fast_sampling_diffusion} and learning based sampling~\cite{watson2022learning_fast_sampling, zheng2022truncated_diffusion, salimans2022progressive_distill_sampling}.
There are also methods like LDM~\cite{rombach2022LDM}, LSGM~\cite{song2020score_generative_SDE} and RDM~\cite{huang2022riemannian_diffusion_model} that utilize distribution learning in the latent space to significantly reduce computational overhead and enable the generation of high-resolution image~\cite{fefferman2016testing_manifold_hypo,yang2022diffusion_model_survey}.
Our I2VGen-XL applies the LDM framework in both two stages in this report.

\textbf{Image synthesis via diffusion models.}
In the current mainstream of diffusion model-based image generation, language models, \eg,  and T5~\cite{raffel2020T5},  are utilized to extract features and cross-attention is employed as a conditioning mechanism to control the generated content.
Typically, Stable diffusion~\cite{rombach2022LDM} and DALL-E 2~\cite{ramesh2022Dalle-2} apply CLIP~\cite{radford2021CLIP} text encoder, and Imagen~\cite{saharia2022Imagen} applies the T5 to improve text-image alignment and image quality.
Following them, some methods aim to perform text-based image editing with diffusion models like Imagic~\cite{kawar2022Imagic} and Sine~\cite{zhang2023sine}.
In addition, another significant research branch is controllable generation, where additional conditions are used to generate target images in a more flexible manner, such as ControlNet~\cite{zhang2023controlnet} and Composer~\cite{huang2023composer}.

\textbf{Video synthesis via diffusion models.}
Early research in video generation~\cite{yu2022video_implicit_GAN,skorokhodov2022stylegan-v,hong2022cogvideo,zhao2018learning} primarily focuses on using GAN-related methods, but maintaining spatio-temporal coherence as well as realism remains a significant challenge.
Inspired by the significant advancements in image generation, diffusion models are also becoming mainstream techniques in video generation~\cite{hoppe2022diffusion_video_infilling,luo2023videofusion,harvey2022flexible_diffusion_video,singer2022make-a-video}.
VDM~\cite{ho2022video_diffusion_models}, VideoLDM~\cite{yang2022DPM_video} and Modelscope-T2V~\cite{wang2023modelscope} achieve significant performance improvement by designing a UNet with temporal-awareness capability to directly generate complete video chunks. 
Instead of modeling the video distribution in the pixels space, MagicVideo~\cite{zhou2022magicvideo} does not use temporal convolution by designing an extra adapter. 
Just like in image generation, controllability is also a crucial objective in the field of video generation~\cite{hu2023videocontrolnet,chen2023control,chu2023video,xing2023make}.
Typically, Gen-1~\cite{esser2023gen-1} utilizes depth as an additional condition to decouple structure and content and achieves excellent results in video-to-video translation tasks.
VideoComposer~\cite{wang2023videocomposer} enables flexible control over video generation by combining textual conditions, spatial conditions, and temporal conditions.
Furthermore, Dragnuwa~\cite{yin2023dragnuwa} takes controllability to the next level by allowing users to control video motion models through simple drag-and-drop gestures, enhancing controllability even further.
Generating high-definition videos has always been an important goal in this field, and significant progress has also been made recently~\cite{zeroscope, ho2022imagenvideo}.
Imagen Video~\cite{ho2022imagenvideo} and Lavie~\cite{wang2023lavie} synthesize high-definition videos to achieve performance improvement in a progressive manner.
As a concurrent method, Videocrafter1~\cite{chen2023videocrafter1,xing2023dynamicrafter} leverages a diffusion model to address high-quality video generation tasks and places significant emphasis on fostering the growth of the community.
Compared to them, I2VGen-XL, as an alternative approach, focuses on enhancing the image-to-video task to enhance video quality, especially in the domain of video content creation, and jointly promotes the advancement of the community.

\section{I2VGen-XL}
\noindent

In this section, we will present a comprehensive introduction of the proposed I2VGen-XL, illustrating how it improves the resolution and spatio-temporal consistency of generated videos while simultaneously preserving the content of input images.
To begin with, we will present a concise summary of the fundamental principles underlying the latent diffusion model.
Subsequently, we delve into the design details and relevant concepts of I2VGen-XL, as shown in Fig.~\ref{fig:framrwork}.
Finally, we will elucidate its training strategy and the inference process.

\subsection{Preliminaries}\label{sec:VLDMs}

Latent Diffusion Model (LDM)~\cite{rombach2022LDM, blattmann2023align_latents} is a kind of effective and efficient diffusion model that gradually recovers the target latent from Gaussian noise, preserving visual manifold, and ultimately reconstructs high-fidelity images or videos from the latent.
For a video $\bm{x} \in \mathbb{R}^{F \times H \times W \times 3}$, we follow VideoComposer~\cite{wang2023videocomposer} to use the encoder of a pretrained VQGAN~\cite{PatrickEsser2021TamingTF} to compress it into a low-dimensional latent representation $\bm{z} = \mathcal{E}(\bm{x})$, where $\bm{z} \in \mathbb{R}^{F \times h \times w \times c}$.
Inversely, the corresponding decoder $\mathcal{D}$ can map the latent representations back to the pixel space $\bar{\bm{x}} = \mathcal{D}(\bm{z})$.

LDM consists of \emph{diffusion} and \emph{denoising} processes.
During the \emph{diffusion} stage, it gradually injects noise to $\bm{z}$ to obtain a noise-corrupted latent $\bm{z}_{t}$, where $t \in \{1,...,T\}$ ($T=1000$ in this work).
During the \emph{denoising} stage, it applies a denoising function $\epsilon_{\theta}(\cdot, \cdot, t)$ on $\bm{z}_{t}$.
The optimized objective can be formulated as follows:
\begin{equation}
    \mathcal{L}_{VLDM} = \mathbb{E}_{\mathcal{E}\bm(x), \epsilon \in \mathcal{N}(0,1), t} \left[\| \epsilon - \epsilon_{\theta}(\bm{z}_{t}, t) \|_{2}^{2}\right].
\end{equation}
Regarding $\epsilon_{\theta}(\cdot, \cdot, t)$, we inherit the well-designed 3D UNet form~\cite{wang2023videocomposer, wang2023modelscope} because of its powerful temporal modeling abilities .
In this report, we will refer to LDM that utilizes a 3D architecture as VLDM unless otherwise specified.

\subsection{I2VGen-XL}\label{sec:i2vgen-xl}

The purpose of I2VGen-XL is to generate high-quality videos from static images.
Therefore, it needs to achieve two crucial objectives: 
\emph{semantic consistency}, which involves accurately predicting the intent in the images and then generating precise motions while maintaining the content as well as the structure of the input images; 
\emph{high spatio-temporal coherence and clarity}, which are fundamental attributes of a video and critical for ensuring the potential of video creation applications.
To this end, I2VGen-XL decomposes these two objectives through a cascaded strategy consisting of two stages: \emph{base stage} and \emph{refinement stage}.

\begin{table}[t]
\centering
\tablestyle{3pt}{1.0}
\begin{tabular}{p{1.2cm}cc}
\shline
 \textbf{Stage} & \textbf{Paramters}  & \textbf{Output Size} \\
\hline
Stem   
        & \( \left[\begin{array}{c} 
            \text{C(4, 64, 5$^\text{2}$, 1, 1)} \\ 
            \text{SiLU()}             \\
            \text{C(64, 64, 3$^2$, 1)} \\
            \text{SiLU()}             \\
        \end{array} \right] \)         \vspace{0.5mm}                              & $\text{32} \times \text{48} \times \text{64}$ \\
\hline
Pool    & A.A.P(32, 32)                                                            & $\text{32} \times \text{32} \times \text{64}$ \\
\hline
Stage 1 
        & \( \left[\begin{array}{c}
            \text{C(64, 256, 3$^2$, 2, 1)} \\
            \text{SiLU()} \\
            \text{C(256, 256, 3$^2$, 1, 1)} \\
            \text{SiLU()} \\
        \end{array} \right] \)    \vspace{0.5mm}                                  & $\text{16} \times \text{16} \times \text{256}$  \\
\hline \vspace{0.5mm} 

Stage 2 
        & \( \left[\begin{array}{c} 
            \text{C(256, 512, 3$^2$, 2, 1)} \\
            \text{SiLU()} \\
            \text{C(512, 512, 3$^2$, 1, 1)} \\
            \text{SiLU()} \\
        \end{array}\right] \)       \vspace{0.5mm}                                 & $\text{8} \times \text{8} \times \text{512}$  \\
\hline
Stage 3 
        & \( \left[\begin{array}{c} 
            \text{C(512, 512, 3$^2$, 2, 1)} \\
            \text{SiLU()} \\
            \text{C(512, 512, 3$^2$, 1, 1)} \\
            \text{SiLU()} \\
        \end{array}\right] \)    \vspace{0.5mm}                                   & $\text{4} \times \text{4} \times \text{512}$  \\
\hline
Stage 4 
        & \( \left[\begin{array}{c} 
            \text{C(512, 512, 3$^2$, 2, 1)} \\
            \text{SiLU()} \\
            \text{C(512, 1024, 3$^2$, 1, 1)} \\
            \text{SiLU()} \\
        \end{array}\right] \)     \vspace{0.5mm}                                   & $\text{2} \times \text{2} \times \text{1024}$  \\
\hline
Output  & C(1024, 1024, 2$^\text{2}$, 2, 1)                                        & $\text{1} \times \text{1} \times \text{1024}$  \\
\shline
\end{tabular}
\caption{
The architecture of the global encoder. 
The `C' represents the Conv2D with the corresponding parameters being input dimension, output dimension, kernel size, stride, and padding.
`A.A.P' is the 2D Adaptive Average Pooling operator.
}
\vspace{-3mm}
\label{tab:cencoder}
\end{table}

\noindent \textbf{Base stage.}
Based on a VLDM, we design the first stage at low resolutions (\emph{i.e.}, $448 \times 256$), which primarily focuses on incorporating multi-level feature extraction, including high-level semantics and low-level details learning, on the input images to ensure intent comprehension and preserve content effectively.

\emph{i)} High-level semantics learning. 
The most straightforward method is to refer to our previous attempt\footnote{https://modelscope.cn/models/damo/Image-to-Video/summary} 
that applies the visual encoder of CLIP to extract the semantic features.
However, we observed that this approach resulted in poor preservation of the content and structure of the input image in the generated videos.
The main reason is that CLIP's training objective is to align visual and language features, which leads to learning high-level semantics but disregards the perception of fine details in the images.
To alleviate this issue, we incorporate an additional trainable global encoder (\emph{i.e}, G.Enc.) to learn complementary features with the same shape, whose architecture is shown in Tab.~\ref{tab:cencoder}.
Then, the two one-dimensional features are integrated via addition operation and embedded into each spatial layer of the 3D UNet using cross-attention.
Despite these efforts, compressing the input images into a low-dimensional vector still leads to information loss.

\emph{ii)} Low-level details. 
To mitigate the loss of details, we employ the features extracted from the encoder of VQGAN (\emph{i.e}, D.Enc.) and directly add them to the input noise on the first frame.
This choice is made based on the encoder's capability to fully reconstruct the original image, ensuring the preservation of all its details.
Our experiments reveal that employing a local encoder instead of a more complex semantic encoder leads to videos with better preservation of image content.
However, as the video plays on, noticeable distortions emerge, suggesting a diminishing clarity of semantics. 
This highlights the complementary nature of the two hierarchical encoders, indicating that their integration is advantageous.

\begin{figure*}[t]
    \includegraphics[width=1.0\linewidth]{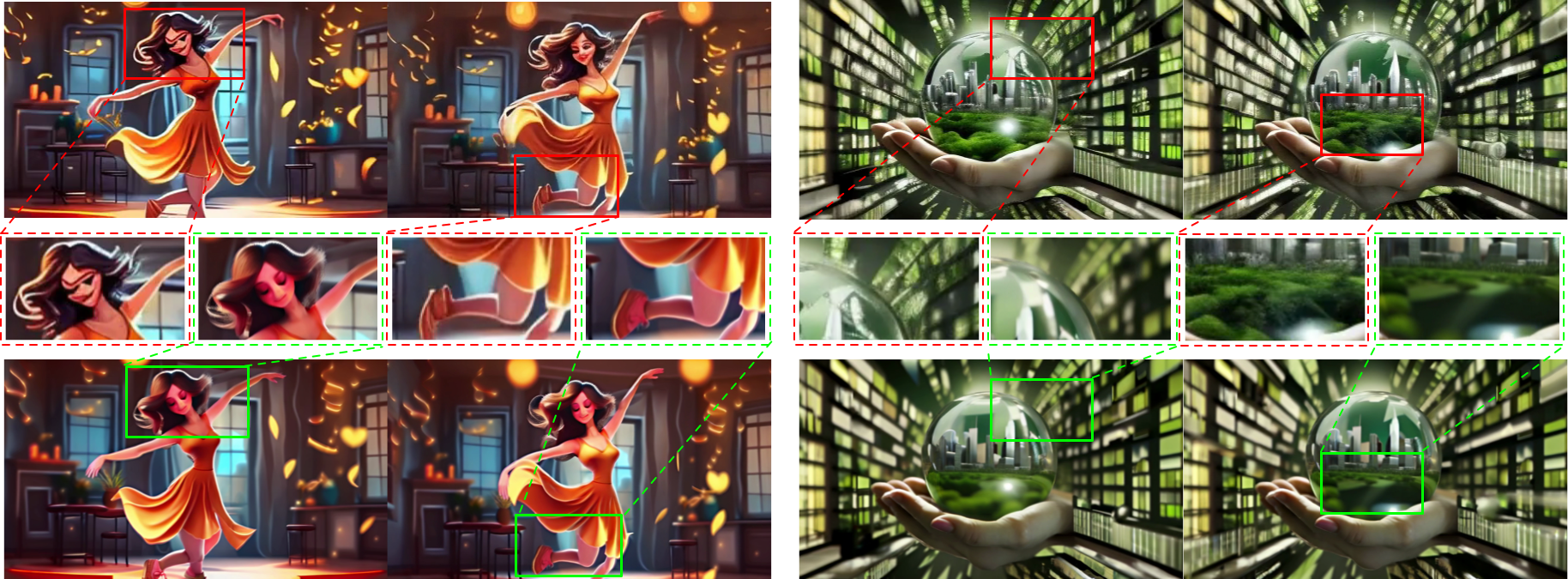}
    \vspace{-6mm}
    \caption{The results of the refinement model. It is evident that the refinement model effectively restores video details and ensures temporal continuity, thus proving its effectiveness.}
    \label{fig:refine_model}
    \vspace{-4mm}
\end{figure*}

\noindent
\textbf{Refinement stage.}
From the base model, we can obtain a low-resolution video with diverse and semantically accurate motions. 
However, these videos may suffer from various issues such as noise, temporal and spatial jitter, and deformations.
Thus the refinement model has two main objectives: \emph{i)} to enhance the video resolution, increasing it from $448 \times 256$ to $1280 \times 720$ or higher; \emph{ii)} to improve the spatio-temporal continuity and clarity of the video, addressing artifacts in both time and space.

To improve the video quality, we train a separate VLDM that specializes in high-quality, high-resolution data and employ a noising-denoising process as introduced by SDEdit~\cite{meng2021sdedit} on the generated videos from the first stage.
Unlike the base model, the refinement model uses the user-provided simple text (\eg a few words) as a condition rather than the original input image.
The reason is that we have found a significant decrease in the effectiveness of video correction when the input conditions for both stages are the same.
This could be due to the introduction of similar mappings with identical conditions, leading to a lack of restorative capability in the model. 
On the other hand, introducing a different condition brings about effective compensation.

Specifically, we encode the text using CLIP and embed it into the 3D UNet by cross-attention. 
Then, based on the pre-trained model from the base stage, we train a high-resolution model using carefully selected high-quality videos, all of which have resolutions greater than $1280\times720$.

\subsection{Traning and Inference}\label{sec:train_inference}

\noindent
\textbf{Training.} 
For the base model, we initialize the spatial component of the 3D UNet with pre-trained parameters from SD2.1~\cite{blattmann2023align_latents}, enabling I2VGen-XL to possess initial spatial generation capabilities.
During the training of the entire 3D UNet, we moderate the parameter updating for the spatial layers by applying a coefficient $\gamma = 0.2$, which is a scaled-down factor compared to that of the temporal layers.

For the refinement model, we initialize it using the well-trained base model and adopt the same training strategy.
To enhance the concentration of the refinement model on spatio-temporal details, we train it specifically on the initial $T_r$ noise scales for denoising.
Furthermore, we adopt a two-stage training strategy:
\emph{i)} We conduct high-resolution training across the entire high-resolution dataset.
\emph{ii)} In order to further enhance the model's perception of details, we carry out a final round of fine-tuning on a meticulously chosen subset of around one million high-quality videos.

\noindent
\textbf{Inference.}
As mentioned above, we employ a noising-denoising process to concatenate the two parts of the model like~\cite{blattmann2023align_latents}.
During inference, we use DDIM~\cite{zhang2022gddim} and DPM-solver++~\cite{lu2022dpm} by considering the generation efficiency and generation quality at different resolutions.
After obtaining the low-resolution videos in the first stage, we resize them to $1280 \times 720$. 
We perform $T_r$ reverse computations with noise addition using DDIM on the new latent space. 
Then, we use the refinement model for the first $T_r$ denoising scales to obtain the final high-resolution videos.
The generative process can be formalized as:
\begin{equation}
    \hat{\epsilon}_{\theta,t}(\bm{z}_{t}, \bm{c}_i, \bm{c}_t, t) = \epsilon_{\theta}(\bm{z}_{t}, \bm{c}_{t}, \hat{\epsilon}_{\theta,i}(\bm{z}_t, \bm{c}_i, T), t) 
    % \epsilon_{\theta}(\bm{z}_{t}, \bm{c}_{1}, t) + \omega \left(\epsilon_{\theta}(\bm{z}_{t}, \bm{c}_{2}, t) - \epsilon_{\theta}(\bm{z}_{t}, \bm{c}_{i}, t)\right),
\end{equation}
where $\bm{c}_i$ and $\bm{c}_t$ represent the input image and text conditions, respectively;
$\hat{\epsilon}_{\theta,t}$ and $\hat{\epsilon}_{\theta, i}$ denote the denoising processes of the base model and the refinement model, respectively;
$T$ is the total number of noise scales in the base stage.
\vspace{-3mm}

\section{Experiments}

\subsection{Experimental Setup}

\noindent
\textbf{Datasets.}
To optimize I2VGen-XL, we use two kinds of training data, \emph{i.e.}, public datasets which comprise  WebVid10M~\cite{2021Frozen} and LAION-400M~\cite{schuhmann2021laion}, as well as private datasets consisting of video-text pairs and image-text pairs of the same type.
In total, these two datasets consist of 35 million videos and 6 billion images, with resolutions ranging from $360p$ to $2k$. 
%
% Additionally, we have standardized the frame rate of all videos to 24 FPS. 
%
We then sorted them based on aesthetic score, motion intensity, and proportion of the main subject. 
This facilitates training with balanced samples.

\begin{figure*}[t]
    \includegraphics[width=1.0\linewidth]{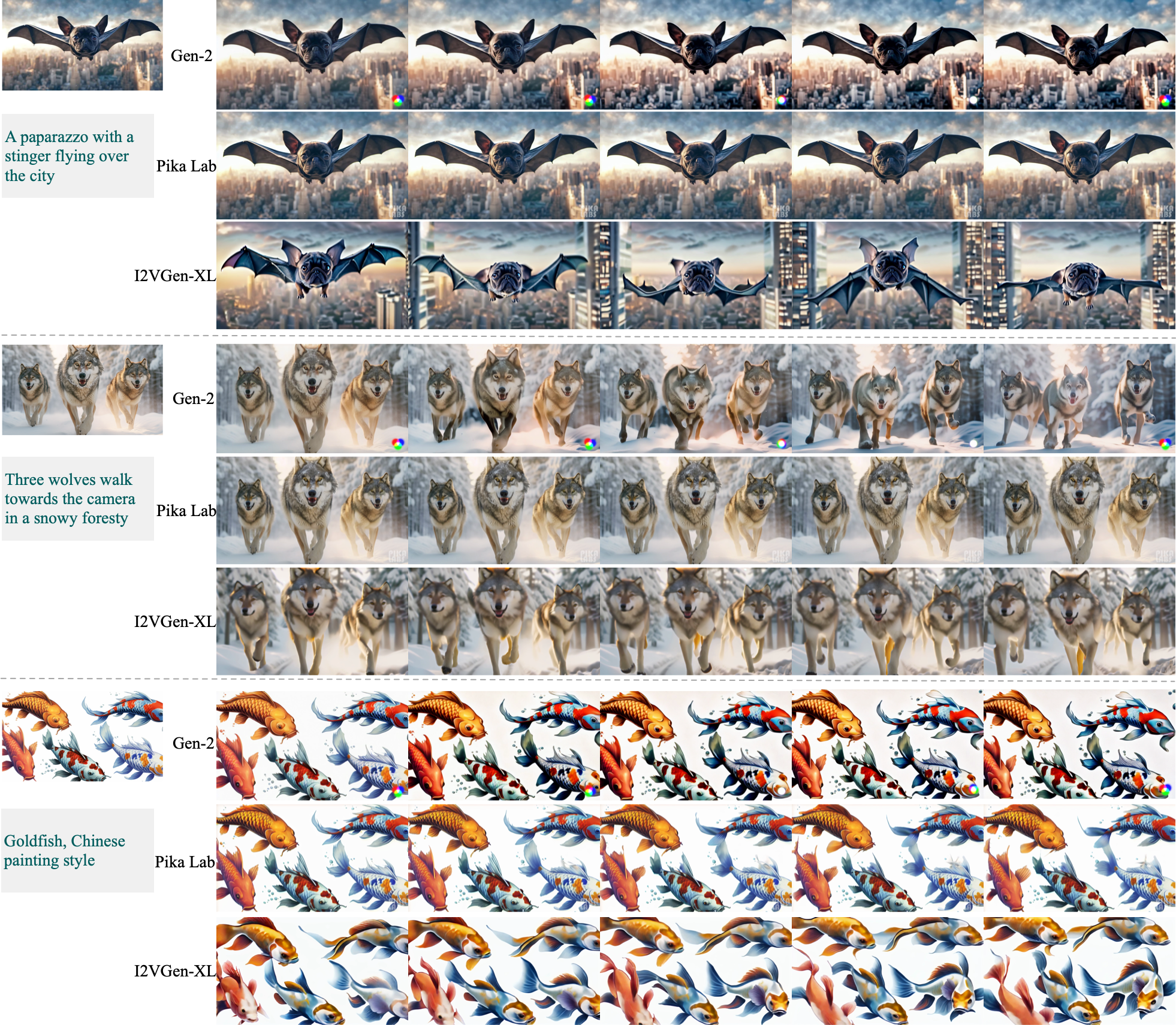}
    \vspace{-6mm}
    \caption{Comparations with Gen-2 and Pika Labs.}
    \label{fig:sota}
    \vspace{-3mm}
\end{figure*}

\noindent
\textbf{Implementations.}
We adopt AdamW~\cite{loshchilov2017AdamW} as the optimizer with a fixed learning rate of $8\times 10^{-5}$.
We simultaneously utilize dynamic frames and dynamic FPS during training. 
For frame length, we use a data supply ratio of $1:1:1:5$ for $1$, $8$, $16$, and $32$ frames respectively. 
Similarly, we use a ratio of $1:2:4:1$ for $1$, $4$, $8$, and $16$ FPS, and it indicates that the input is a static image when FPS is equal to $1$.
We use center crop to compose the input videos with $H = 256, W = 448$ and $H = 720, W = 1280$ for the base and refinement stages respectively.
For training the diffusion models, we employ the v-parameterization of diffusion models and offset noise with a strength of $0.1$. 
The linear schedule is used.
During inference, the default value of $T_r$ is set to $600$, but it may vary for certain examples.

\subsection{Experimental Results}

\noindent
\textbf{Compartion with top methods.}
To demonstrate the effectiveness of our proposed method, we compare the performance of I2VGen-XL with the leading methods Gen-2~\cite{esser2023gen-1} and Pika~\cite{pikalab}, which are widely recognized as the current state-of-the-arts in the field.
We used their web interfaces to generate videos with three types of images, including pseudo-factual, real, and abstract paintings, as shown Fig.~\ref{fig:sota}.
From these results, several conclusions can be drawn:
\emph{i)} Richness of motions: our results exhibit more realistic and diverse motions, such as the example of a flying dog. 
In contrast, the videos generated by Gen-2 and Pika appear to be closer to a static state, indicating that I2VGen-XL achieves a better richness of motions;
\emph{ii)} Degree of ID preservation: from these three samples, it can be observed that Gen-2 and Pika successfully preserve the identity of the objects, while our method loses some details of the input images. 
In our experiments, we also found that the degree of ID preservation and the intensity of motion exhibit a certain trade-off relationship. 
We strike a balance between these two factors.

\begin{figure*}[t]
    \includegraphics[width=1.0\linewidth]{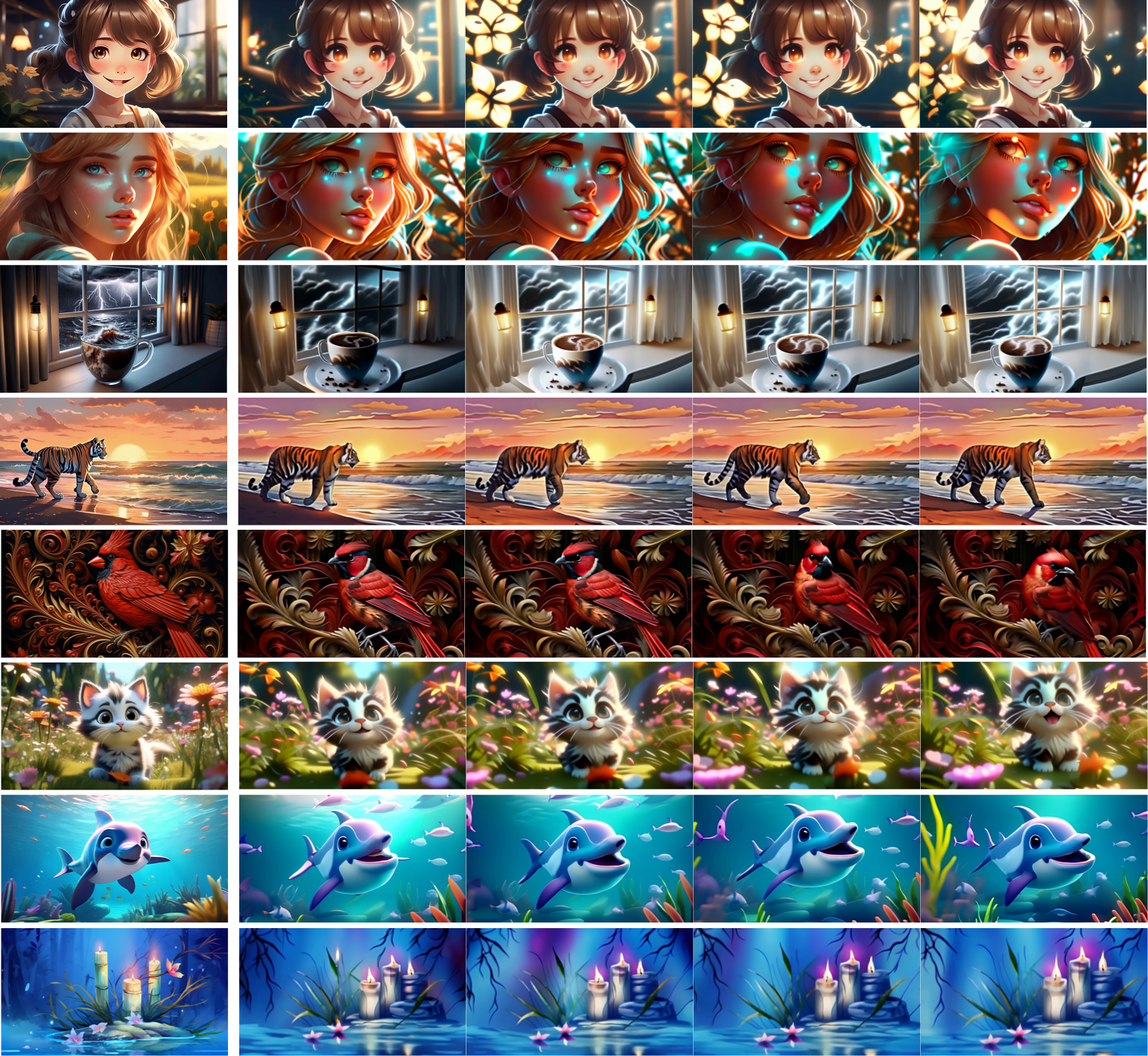}
    \vspace{-6mm}
    \caption{More videos from diverse categories, including people, animals, anime, Chinese paintings, and more.}
    \label{fig:more_results}
    \vspace{-3mm}
\end{figure*}

\begin{figure*}[h]
    \includegraphics[width=1.0\linewidth]{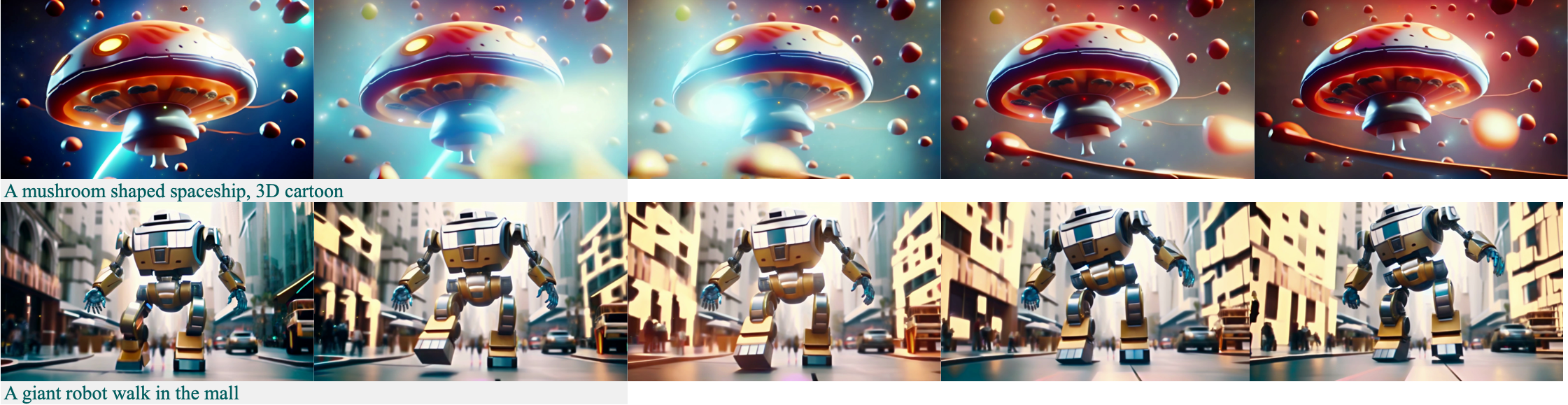}
    \vspace{-6mm}
    \caption{Generated videos from input text prompts.}
    \label{fig:t2v}
    \vspace{-3mm}
\end{figure*}

\begin{figure}[t]
    \centering
    \subfloat{
	\begin{minipage}[t]{0.47\linewidth}
		\centering
		\includegraphics[width=1\linewidth]{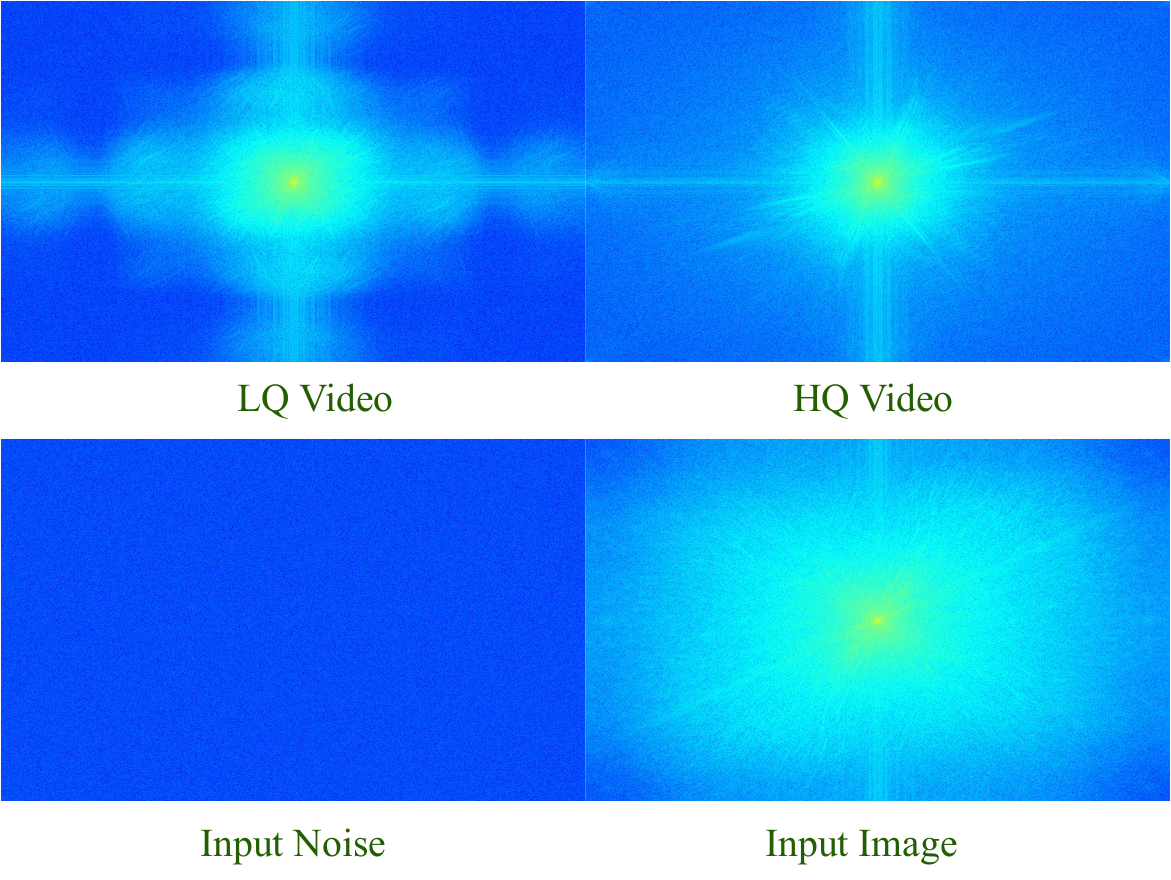}
            \subcaption{\scriptsize{Spatial spectrogram}}
            \label{fig:01_spatial_spectrogram}
	\end{minipage}
        \vspace{5mm}
        \begin{minipage}[t]{0.50\linewidth}
		\centering
		\includegraphics[width=1\linewidth]{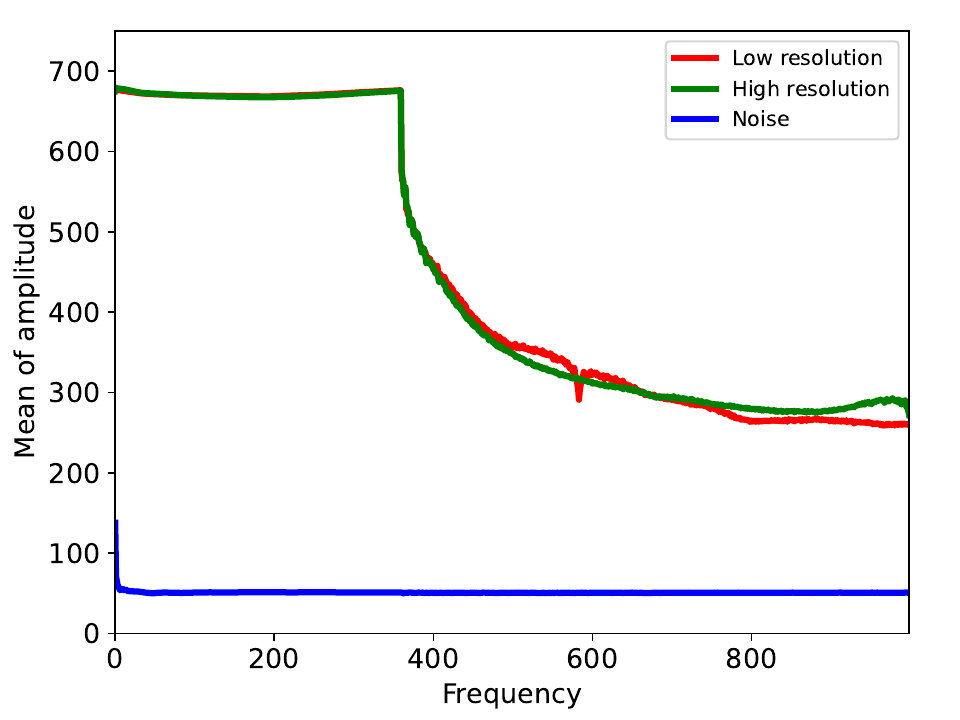}
            \subcaption{\scriptsize{Spatial  distribution}}
            \label{fig:02_spatial}
	\end{minipage}
    }
    \vspace{-3mm}
    \subfloat{
	\begin{minipage}[t]{0.47\linewidth}
		\centering
		\includegraphics[width=1\linewidth]{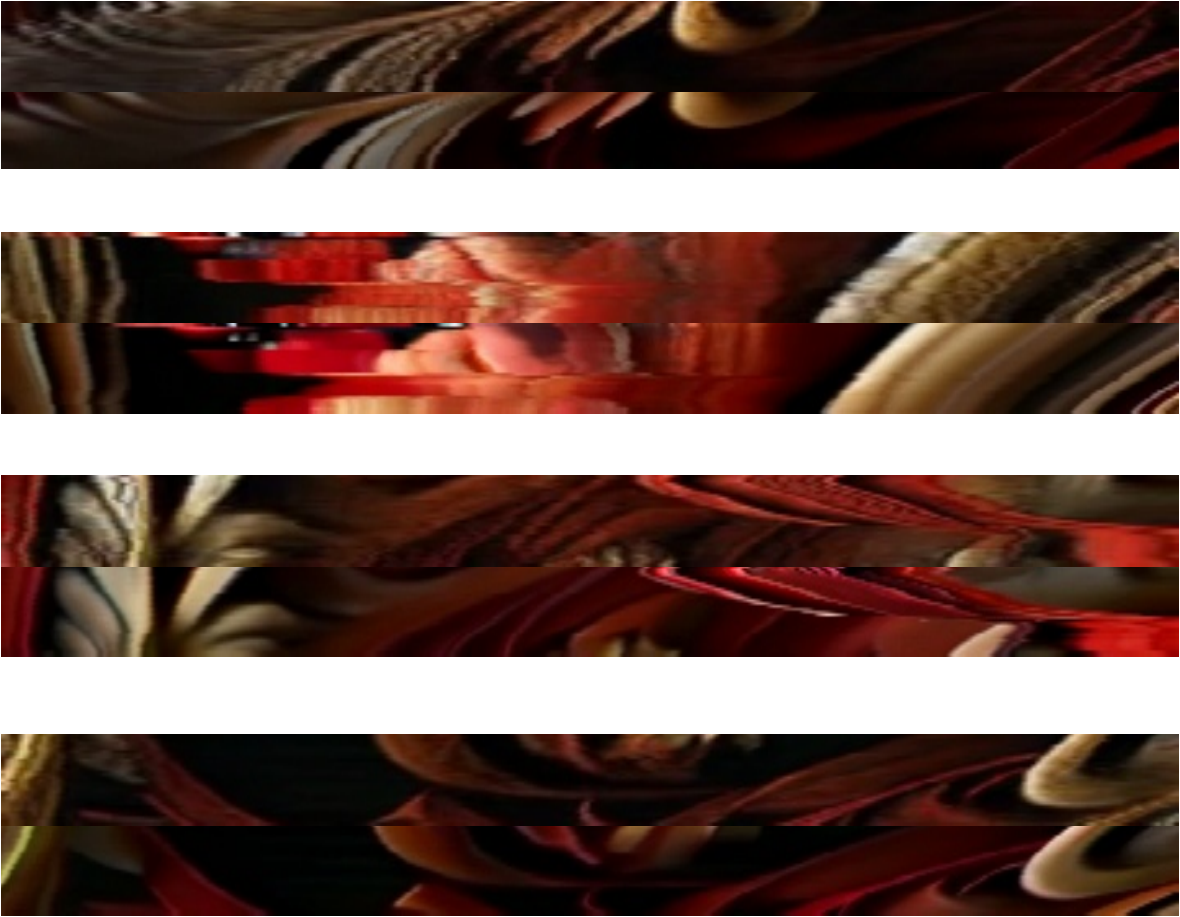}
            \subcaption{\scriptsize{Temporal section}}
            \label{fig:03_temporal_section}
	\end{minipage}
        \begin{minipage}[t]{0.50\linewidth}
		\centering
		\includegraphics[width=1\linewidth]{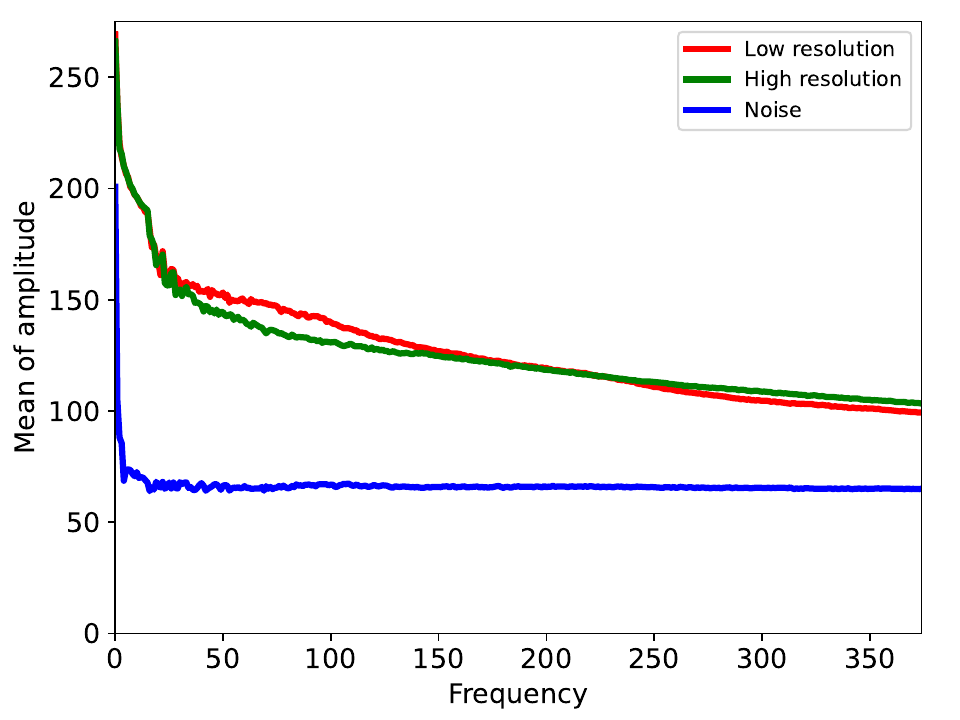}
            \subcaption{\scriptsize{Temporal distribution}}
            \label{fig:04_temporal_spect_0100}
	\end{minipage}
    }
    \vspace{-2mm}
    \caption{Frequency spectrum analysis of the refinement model.}
    \label{fig:st_continuity}
    \vspace{-3mm}
    \label{fig:}
\end{figure}

\noindent
\textbf{Analyse of Refinement model.}
Fig.~\ref{fig:refine_model} illustrates the generated videos before and after the refinement stage. 
These results reveal substantial enhancements in spatial details, including refined facial and bodily features, along with a conspicuous reduction in noise within local details.
To further elucidate the working mechanism of the refinement model, we analyzed the spatial and temporal changes occurring in the generated videos during this process in the frequency domain in Fig.~\ref{fig:st_continuity}.
Fig.~\ref{fig:01_spatial_spectrogram} presents the frequency spectra of four spatial inputs, which reveals that low-quality video exhibits a frequency distribution similar to that of noise in the high-frequency range, while high-quality videos demonstrate a closer resemblance to the frequency distribution of the input image. 
Combining this with the spatial frequency distribution depicted in Fig.~\ref{fig:02_spatial}, it can be observed that the refinement model effectively preserves low-frequency data while exhibiting smoother variations in high-frequency data.
From the perspective of the temporal dimension, Fig.~\ref{fig:03_temporal_section} presents the temporal profiles of the low-quality video (top) and the high-quality video (bottom), demonstrating a clear improvement in the continuity of the high-definition video. 
Additionally, combining Fig.~\ref{fig:02_spatial} and Fig.~\ref{fig:04_temporal_spect_0100}, it can be observed that the refinement model preserves low-frequency components, reduces mid-frequency components, and enhances high-frequency components, both in the spatial and temporal domains. 
This indicates that the artifacts in the spatio-temporal domain mainly reside in the mid-frequency range.

\noindent
\textbf{Qualitative analysis.}
We also conducted experiments on a wider range of images, including categories such as human faces, 3D cartoons, anime, Chinese paintings, and small animals. 
The results are shown in Fig.~\ref{fig:more_results}.
We can observe that the generated videos simultaneously consider the content of the images and the aesthetics of the synthesized videos, while also exhibiting meaningful and accurate motions. 
For example, in the sixth row, the model accurately captures the cute mouth movements of the kitten. 
Additionally, in the fifth row, the woodblock bird model accurately rotates its head while maintaining the original style. 
These results indicate that I2VGen-XL exhibits promising generalization ability.

\begin{figure}[t]
    \includegraphics[width=1.0\linewidth]{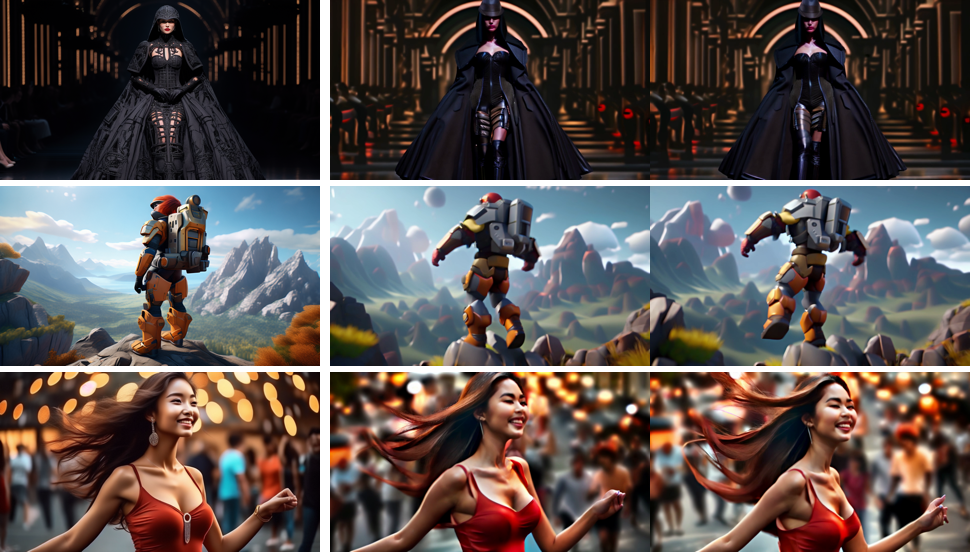}
    \vspace{-6mm}
    \caption{Generated videos of human bodies by I2VGen-XL.}
    \label{fig:person}
    \vspace{-3mm}
\end{figure}

% \noindent
% \textbf{Human-Specific Models.}
% %
Generating stable human body movements remains a major challenge in video synthesis.
Therefore, we also specifically validated the robustness of I2VGen-XL on human body images, as shown in Fig.~\ref{fig:person}. 
It can be observed that the model's predictions and generated motions for human bodies are reasonably realistic, with most of the characteristic features of human bodies well-preserved.

\noindent
\textbf{Text-to-video.}
One of the main challenges in text-to-video synthesis currently is the collection of high-quality video-text pairs, which makes it more difficult to achieve semantic alignment between the video and text compared to image synthesis. 
Therefore, combining image synthesis techniques, such as Stable Diffusion~\cite{blattmann2023align_latents} and Wanxiang\footnote{https://wanxiang.aliyun-inc.com}, with image-to-video synthesis can help improve the quality of generated videos. 
In fact, to respect privacy, almost all samples in this report are generated by combining these two. 
Additionally, in Fig.~\ref{fig:t2v}, we separately generate samples, and it can be observed that the video and text exhibit high semantic consistency.

\noindent
\textbf{Limitations.} 
The proposed I2VGen-XL in this report has shown significant improvements in both semantic consistency and spatio-temporal quality in video synthesis. 
However, it still has the following limitations, which also serve as our future works:
\emph{i) Human body motion generation}, as mentioned earlier, still presents significant challenges in terms of freedom and naturalness. 
This is primarily due to the intricate and rich nature of human-related movements, which increases the difficulty of generation;
\emph{ii) Limited ability to generate long videos.} Current models mainly generate short videos of a few seconds with a single shot, and they are not yet capable of generating long videos with continuous storytelling and multiple scenes;
\emph{iii) Limited user intent understanding.} The current scarcity of video-text paired data restricts the ability of video synthesis models to comprehend user inputs, such as captions or images, effectively. 
This significantly increases the difficulty for users to interact with the models.

\section{Conclusion}
In this report, we introduced a cascaded video synthesis model called I2VGen-XL, which is capable of generating high-quality videos from a single static image. 
We approached the problem from two perspectives: semantic consistency and spatio-temporal continuity, and devised two stages to address each of these two purposes separately. 
The effectiveness of the I2VGen-XL method is validated using a substantial amount of category data. 
Additionally, we explored a new paradigm of video synthesis by combining the I2VGen-XL approach with the image synthesis algorithm.
However, despite these efforts, we know that video synthesis still faces numerous challenges. 
%
%To approach the level of effectiveness seen in current image synthesis, 
Further exploration is needed in areas such as human objects, duration, and intent understanding, in order to address more practical tasks in video synthesis.

{\small
\bibliographystyle{ieee_fullname}
\bibliography{egbib}
}

\end{document}